\documentclass[10pt]{article}

\usepackage[T1]{fontenc}        %
\usepackage{dialogue}           %
\usepackage{tabularx}           %
\usepackage{amssymb}            %
\usepackage{amsmath}            %
\usepackage{algorithm}          %
\usepackage{algpseudocode}      %
\usepackage{float}              %
\usepackage{mdframed}           %
\usepackage{xcolor}             %
\usepackage{threeparttable}     %
\usepackage{graphicx}           %
\usepackage{enumitem}           %
\usepackage{authblk}            %
\usepackage{url}

\title{Can Large Language Models Match Tutoring System Adaptivity? A Benchmarking Study}

\author{Conrad Borchers}
\author{Tianze Shou}

\affil{Carnegie Mellon University}
\affil{\texttt{\{cborcher,tshou\}@cs.cmu.edu}}

\date{} %

\begin{document}

\maketitle

\begin{abstract}
Large Language Models (LLMs) hold promise as dynamic instructional aids. Yet, it remains unclear whether LLMs can replicate the adaptivity of intelligent tutoring systems (ITS)—where student knowledge and pedagogical strategies are explicitly modeled. We propose a prompt variation framework to assess LLM-generated instructional moves' adaptivity and pedagogical soundness across 75 real-world tutoring scenarios from an ITS. We systematically remove key context components (e.g., student errors and knowledge components) from prompts to create variations of each scenario. Three representative LLMs (Llama3-8B, Llama3-70B, and GPT-4o) generate 1,350 instructional moves. We use text embeddings and randomization tests to measure how the omission of each context feature impacts the LLMs' outputs (adaptivity) and a validated tutor-training classifier to evaluate response quality (pedagogical soundness). Surprisingly, even the best-performing model only marginally mimics the adaptivity of ITS. Specifically, Llama3-70B demonstrates statistically significant adaptivity to student errors. Although Llama3-8B's recommendations receive higher pedagogical soundness scores than the other models, it struggles with instruction-following behaviors, including output formatting. By contrast, GPT-4o reliably adheres to instructions but tends to provide overly direct feedback that diverges from effective tutoring, prompting learners with open-ended questions to gauge knowledge. Given these results, we discuss how current LLM-based tutoring is unlikely to produce learning benefits rivaling known-to-be-effective ITS tutoring. Through our open-source benchmarking code, we contribute a reproducible method for evaluating LLMs' instructional adaptivity and fidelity.
\end{abstract}

\section{Introduction and Related Work} 

Recent advances in large language models (LLMs) have sparked interest in their potential to enhance (or replace) intelligent tutoring systems (ITS) and other adaptive learning systems by providing real-time, conversational support to learners. ITS rely on rule-based models to guide students through problem-solving processes, leveraging domain knowledge derived from cognitive task analysis, learner modeling, and pedagogical strategies known to enhance learning \cite{vanlehn2006behavior,koedinger2012knowledge,huang2021general}. In contrast, LLMs generate responses based on statistical patterns in language rather than explicit instructional logic \cite{liffiton2023codehelp,stamper2024enhancing}. While advancements have been made to integrate instructional principles into LLMs through prompt engineering \cite{liffiton2023codehelp,stamper2024enhancing,venugopalan2024combining}, this contrast raises critical questions about whether LLMs can maintain pedagogical coherence by generating instruction aligning with evidence-based principles, such as prompting for self-explanation \cite{bisra2018inducing}.

Whether LLMs can provide instruction similar to ITS is relevant because they are increasingly used in emerging AIED learning environments. For example, hybrid tutoring, which integrates human and AI learning support \cite{thomas2023tutor}, has been proposed as a promising paradigm to enhance student learning experiences with LLMs \cite{venugopalan2024combining}. As the field of AIED increasingly moves toward such human-AI hybrid adaptivity settings, conversational support for tutors and learners through instructional move recommendations is emerging as a key LLM application in AIED \cite{venugopalan2024combining,borchers2024combining}. In this paradigm, LLMs provide real-time scaffolding, tutor-like explanations, and conversational interventions tailored to a tutor's or student's needs. However, while LLMs have demonstrated fluency in natural language generation to provide dialog-based instructional moves, their ability to deliver contextually appropriate guidance has been questioned \cite{stamper2024enhancing,venugopalan2024combining}. Specifically, past research highlighted LLM's limitations in representations of learner knowledge and instruction on specific skills (though they demonstrate some potential in tracing knowledge \cite{scarlatos2024exploring,zhang2024predicting}). Therefore, in addition to pedagogical coherence, we study if LLMs can generate responses that exhibit the structured adaptivity of ITS, addressing contextual relevance.

Despite the growing enthusiasm for integrating LLMs into AIED systems, the field lacks evaluation methods for assessing their effectiveness in providing adaptive support. Exceptions like Karumbaiah et al. \cite{karumbaiah2024evaluating} introduced methods to evaluate LLMs' adherence to pedagogical strategies; yet, these approaches fall short in addressing adaptivity (e.g., by adding learner behavior into prompt instructions during learning \cite{venugopalan2024combining})—an essential feature of ITS. Similarly, emerging work on knowledge tracing with LLMs \cite{zhang2024predicting,scarlatos2024exploring} offers insights into tracking student performance but does not assess how LLMs adjust their responses based on learner progress. This gap in evaluation methods poses an important challenge: without methods to systematically determine whether LLMs can replicate the adaptivity typical for ITS, their integration into hybrid tutoring environments risks being pedagogically ineffective. We investigate the nature of LLM-generated responses in tutoring contexts by investigating the following research questions:
\begin{itemize}
    \item \textbf{RQ1:} Do LLMs respond to adaptivity typical for tutoring systems?
    \item \textbf{RQ2:} Do they do so in a desirable way?
    \item \textbf{RQ3:} What is the diversity and type of generations LLMs provide in the context of hybrid tutoring message recommendations?
\end{itemize}
We contribute bridges between ITS adaptivity and the generative capabilities of LLMs by analyzing how LLMs respond to tutoring scenarios that require dynamic, structured guidance. By investigating the alignment of LLM instruction with best tutoring practices, we contribute open-source methods and code for evaluating the instructional effectiveness of LLMs pre-deployment.\footnote{\url{https://github.com/conradborchers/llm-instruction-benchmarking}}

\section{Methods}

\subsection{Data Set and Study Context}

We collected a dataset from the open-source intelligent tutoring system (ITS) \textit{Lynnette} \cite{long2018exactly}, designed for practicing mathematical equation solving. \textit{Lynnette} is a step-based problem-solving system that guides students through individual steps in solving linear equations, providing immediate feedback on correctness. Students can also request hints. The system employs an underlying skill model that maps each problem-solving step to one or more skills (e.g., "distribute-division"), which we refer to as \textit{knowledge components} (KC) \cite{long2018exactly}.

The dataset, drawn from prior research \cite{venugopalan2024combining}, includes dialogue data between student solvers and their parents. These parents participated in a pilot study testing a conversational tutoring system designed to support their child's engagement with \textit{Lynnette} in an in-person prototyping study. The dataset consists of 10 student-parent dyads, with students working through equation-solving tasks while parents provided guidance and motivation. It includes 75 tutoring scenarios, represented as 30-second log data snippets capturing various interactions, such as students correctly progressing, making mistakes, or engaging in ongoing conversations with their parents. Participants were recruited through a university-affiliated outreach program and social media.

\subsection{Problem-Solving Context and LLM Prompting}

To provide ITS-sourced, real-time information for instructional adaptivity, we define a \textit{problem-solving context} at the prompt engineering stage. This context includes details on the student's progress and any chat-based interactions with the human tutor (i.e., parent). Specifically, we track the \textit{current problem} (e.g., "3(2x + 4) - 2 = 16"), \textit{correct student steps} (e.g., ["3(2x + 4) = 18", "2x + 4 = 6"]), \textit{incorrect steps}, \textit{ITS hints} (if any, e.g., ["How can you get rid of 10 on the right?"]), and the \textit{ITS-suggested next step} (e.g., ["Divide by 2 on both sides: 2x / 2 + 4 / 2 = 6 / 2"]). \textit{Lynnette} 's instructional model enables both the LLM and the parent to view suggested next steps. Additionally, we track \textit{student-parent chat history} (e.g., ["Student: can you help explain why this is incorrect?", "Parent: you are missing division on constant 4"]) and the \textit{knowledge components (KCs)} involved in the current step (e.g., ["divide-const", "distribute-multiplication"]). All problem-solving context components, except for the current problem, dynamically update as the student progresses. 

\begin{figure*} 
    \centering
    \includegraphics[width=1\linewidth]{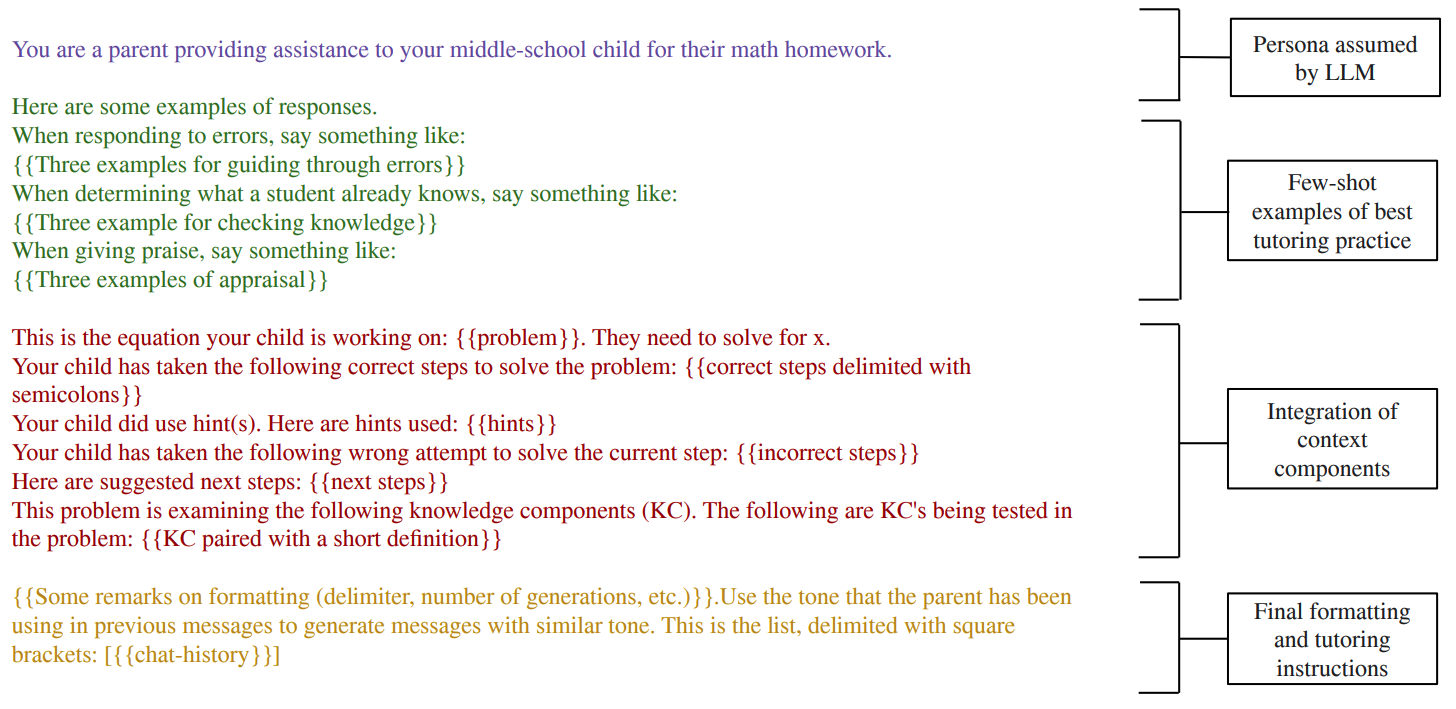}
    \caption{Prompt template for LLM prompt in this study. The red section and explanatory text are variable concerning each problem-solving context.}
    \label{fig:prompt}
\end{figure*}

All problem-solving context components are dynamically incorporated into LLM prompts using a prompt template and \verb|{placeholders}|. Fig. \ref{fig:prompt} illustrates the full prompt sent to the LLMs, where we employ techniques such as persona-based prompting and few-shot learning, dynamically integrating context components into the prompt.

\subsection{Experiment Data Pipeline}

\begin{figure}[]
    \centering
    \includegraphics[width=0.9\textwidth]{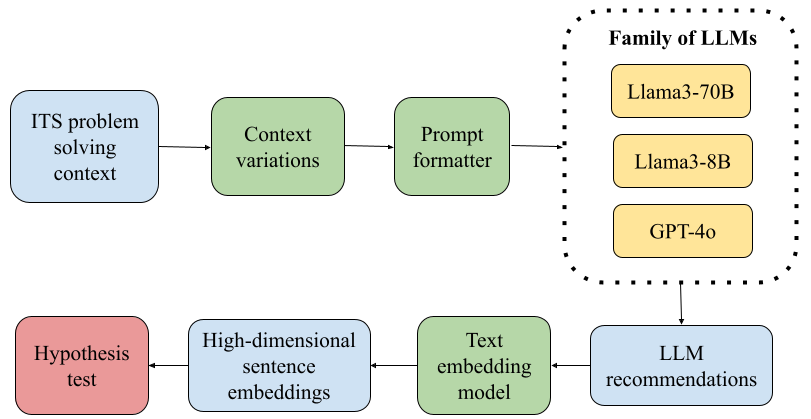}
    \caption{Data pipeline of our experiment. In this diagram, blue components represent data and green components represent data transformations}
    \label{fig:pipeline}
\end{figure}

RQ1 evaluates the responsiveness of LLM-generated recommendations as problem-solving contexts shift. Specifically, the LLM should adapt its guidance based on whether the student solves a step correctly or incorrectly and adjust its response based on different types of errors. RQ2 builds on this by assessing whether these adaptations align with sound pedagogical principles. For effective tutoring, conversational feedback should acknowledge effort, address errors indirectly, and accurately determine student understanding \cite{thomas2023tutor}. We propose an experimental data pipeline to evaluate our LLM system and compare different models on these RQs (Fig. \ref{fig:pipeline}). The pipeline takes problem-solving context examples from the ITS, systematically modifies these contexts using learner data, constructs prompts, and feeds them into the target LLMs. The generated responses are then transformed into text embeddings for further analysis to test whether LLMs adaptively respond to the prompt permutations (see Section \ref{sec:testing}).

To generate context variations (first green box, top row; Fig. \ref{fig:pipeline}), we remove specific components from the problem-solving context to assess LLM responsiveness to ITS adaptivity (RQ1). This process generates modified versions of the 75 scenarios to compare how the inclusion or exclusion of individual components influences LLM recommendations. We create five modified copies of each original context, omitting the following elements: (1) student's correct step history, (2) student's incorrect step history, (3) ITS-suggested next step, (4) KC(s) involved in the current step, and (5) displayed hint(s). This results in 75 context groups, each containing six contexts (one original and five variants). All contexts are then formatted into prompts and fed into three selected LLMs: Llama3-8B, Llama3-70B \cite{grattafiori2024llama3herdmodels}, and GPT-4o \cite{openai2024gpt4ocard}. These models were chosen to represent key archetypes in the LLM ecosystem: (1) a small, cost-effective distilled model (Llama3-8B), which can run locally on standard PCs, (2) a mid-sized open-source model (Llama3-70B), which, as of Fall/Winter 2024, provides competitive performance to state-of-the-art LLMs for tasks such as question-answering, math, and coding \cite{grattafiori2024llama3herdmodels}, and (3) a proprietary, state-of-the-art model (GPT-4o-2024-11-20). Each model generates responses for $75 \times 6$ contexts, yielding $75 \times 6 \times 3$ responses (1,350 total).

\subsection{LLM Recommendation Quality}
\label{sec:recommendation-quality}

We evaluate the pedagogical quality for LLM-generated recommendations (RQ2) through a classifier by Thomas et al. \cite{thomas2024tutors}. The classifier, designed for rating scenario-based tutor training conversations based on evidence-based principles, provides feedback on whether the instructional move (e.g., open-response text) is pedagogically sound (1 if "sound" and 0 otherwise). The classifier achieved high accuracy on human-labeled data with $F_1 \approx 0.8$. The classifier determines if tutoring guidance appropriately praises correct attempts and offers indirect corrections for errors, making it well-suited for our study, where a parent or tutor guides a student problem solver. We also assess the instruction-following ability of LLMs by evaluating adherence to prompt constraints:

\begin{itemize}[leftmargin=0pt] \item[] \textbf{Intention inclusion}: Checks whether the LLM response includes an intention clause, such as \texttt{[Encourage child to continue]} or \texttt{[Correct student's mistake]} for explainability, formatted within brackets as instructed.
\item[] \textbf{Existence of response delimiter}: Verifies inclusion of the delimiter (\#).
\item[] \textbf{Generation of exactly three recommendations}: Ensures that three recommendations are generated, contingent on meeting the delimiter criterion.
\end{itemize}

\subsection{Statistical Testing}
\label{sec:testing}

We assess LLMs' adaptivity (RQ1) to problem-solving contexts through a novel hypothesis testing procedure based on randomization tests. We leverage an encoder-based text embedding model to map textual data into high-dimensional vectors \cite{nie2024textembeddingmeetslarge}. These vectors encode semantic differences in LLM-generated instructional moves. We use these differences to determine whether LLM moves are significantly correlated with learner data to which the LLM should adapt.

The embedded vectors retain two vital properties: 1) two sentences with similar semantic meanings produce embeddings that are in close neighborhood in high-dimensional space; 2) The relative positions of embedding vectors also encode semantic meanings. For example, since the word "dog" and "puppy" have similar meaning, the distance between $\mathbf{v}_{dog}$ and $\mathbf{v}_{puppy}$ should be much smaller than that between $\mathbf{v}_{dog}$ and $\mathbf{v}_{human}$ (i.e. $\|\mathbf{v}_{dog} - \mathbf{v}_{puppy}\|_2 << \|\mathbf{v}_{dog} - \mathbf{v}_{human}\|_2$). As an example of property two, the embeddings among words "king," "queen," "man," and "women" should have the following relative positional relationship: $\mathbf{v}_{queen} \approx \mathbf{v}_{king} - \mathbf{v}_{man} + \mathbf{v}_{woman}$. We rely on these properties by examining the relative position shifts of the LLM generations' embeddings when some components in problem-solving contexts are removed or remain intact. We selected OpenAI's \texttt{text-embedding-3-large} model to transform the LLM output into 3072-dimensional vectors for this experiment.

We assess LLM adaptivity to contextual ITS information (e.g., correct attempts) in the prompt. If an LLM adapts to information, omitting it from the prompt should influence its output. Mathematically, given a matrix $M \in \mathbb{R}^{75 \times 3072}$ (where 75 represents the sample size and 3072 the embedding dimension), we test whether the distributions of variant embeddings differ from those generated by the unmodified prompt. Formally, if sets of embeddings $x_1, x_2$ are sampled from distributions $D_0$ and $D_1$, corresponding to embeddings with and without context information, we propose a hypothesis pair: $H_0$ states embeddings remain invariant to context information, while $H_1$ states they differ. We use approximately ($\approx$) because LLMs will generally generate different, though similar, content when prompted with the same prompt multiple times \cite{cheng2024relic}.
\begin{equation*}
    H_0: f(x_1 | D_0) \approx f(x_2 | D_1);\quad H_1: f(x_1 | D_0) \not\approx f(x_2 | D_1)
\end{equation*}
To statistically test this hypothesis pair, we use distance metrics to capture the average similarity between groups of LLM generations. Two distributions are considered different if the distance between their samples exceeds the expected distance between samples from the same distribution, which arises due to chance \cite{cheng2024relic}. These samples consist of LLM generations from 75 problem-solving context prompts. We determine that two distributions ($X$ and $Y$) are different if:
\begin{equation}
    \mathbb{E}_{X, Y}[d(x, y)] > \mathbb{E}_{X, Y}[d(z, z')]
    \label{eq:expectation-ineq}
\end{equation}
\begin{equation}
    dist_{cos}(\mathbf{u}, \mathbf{v}) = \frac{\mathbf{u} \cdot \mathbf{v}}{\|\mathbf{u}\|_2 * \|\mathbf{v}\|_2}
    \label{eq:cosine-distance}
\end{equation}
where $x \sim X$, $y \sim Y$, and $z$, $z'$ are sampled from $X$ and $Y$ with equal probability (i.e., $z, z' \sim \frac{1}{2}X + \frac{1}{2}Y$), and function $dist$ being the cosine similarity distance metric \cite{cosine-distance-paper} defined in equation \ref{eq:cosine-distance} where $\cdot$ represents the dot product, and $\|*\|_2$ the L2 norm. We compute the left-hand side of inequality \ref{eq:expectation-ineq} using the average distance between embedding pairs produced by the prompt with and without the context information. The average distance $M$ between $M^0$ and $M^i$ is defined as the arithmetic mean of similarity distances ($M = \frac{1}{n} \sum_{j=0}^{n} dist(M^0_j, M^i_j)$).

Focusing on the right-hand side of inequality \ref{eq:expectation-ineq}, to simulate 50-50 sampling from both $X$ and $Y$, we conduct randomized bootstrapping to approximate this expectation. Again, we take $M^0$ and $M^1$ as example in the place of $X$ and $Y$ and demonstrate using the following pseudo-code:
\begin{algorithm}
\caption{Algorithm to approximate the similarity distribution at chance}
\begin{algorithmic}
\State \textbf{Input:} $M^0, M^1 \in \mathbb{R}^{75 \times 3072}$; \textbf{Output:} A vector of length $B$
\State $\tilde{M} = \text{concat}(M^0, M^1)$; output = [\,]
\For{$b \in \{1,2,\dots,B\}$}
    \State $\tilde{M} = \text{shuffle}(\tilde{M})$
    \State $M^a = \tilde{M}[0:75]$; $M^b = \tilde{M}[75:150]$
    \State MeanDist = $\frac{1}{75} \sum_{j=0}^{75} d(M^a_j, M^b_j)$
    \State output.append(MeanDist)
\EndFor; Return output
\end{algorithmic}
\end{algorithm}
As a result, we obtain a distribution of bootstrapped distances of length $B$, where each value serves as a bootstrapped simulated sample from $d(z, z')$ and $z, z' \sim \frac{1}{2}X + \frac{1}{2}Y$, giving us the right-hand side's distribution of inequality \ref{eq:expectation-ineq}. We then use the value obtained from the average distribution distance $M$ as the test statistics and compute the $p$-value using the test statistics' quantile on the simulated distribution. In our experiment, we set $B$ to be 1000. We also computed and reported the effect size of these tests. Formally, we use the Cohen's $d$ effect size given by:
\begin{equation}
    d = \frac{dist(x, y) - \mathbb{E}[dist(z, z')]}{\sqrt{\mathbb{V}[dist(z, z')]}}
\end{equation}
where $dist(x,y)$ denotes the test statistic, and $\mathbb{E}[dist(z, z')]$ and $\sqrt{\mathbb{V}[dist(z, z')]}$ represent the mean and standard deviation of the bootstrap distribution, respectively. A larger effect size indicates a greater divergence between distributions $X$ and $Y$, while a negative effect size suggests that the observed mean of $dist(x,y)$ is smaller than the average distance obtained via random shuffling. Here, a Cohen's $d$ of about 1.96 $SD$ d aligns with 95\% confidence and $p$ = .05.

\subsection{Qualitative Analysis}
\label{sec:qualitative-analysis}

Qualitatively assessing LLM generations (RQ3), we ensure outputs possess face validity beyond the quantitative checks mentioned above (e.g., "Do these LLMs exhibit distinct styles?" and "Do certain LLMs retain characteristics absent in others?"). We also checked for hallucinations and math errors \cite{ji2023survey}. We visualize high-dimensional response embeddings using Principal Component Analysis (PCA) to reduce them to 2D for visualization to discover clusters. The first two principal components captured 24.9\% of the total variance.

\section{Results}

\subsection{RQ1: Can LLMs Match ITS Adaptivity?}

Table \ref{tab:p-values} reveals that no LLMs, except Llama3-70B, exhibit significant responsiveness to context components. Specifically, Llama3-70 B's outputs change significantly when the incorrect steps component is removed ($p = .035$, indicating its influence on the model's responses. No other components show evidence of impact. Positive effect sizes suggest some degree of shift beyond random chance. Although not statistically significant, small distribution differences appear for GPT-4o and Llama3-70B when incorrect steps and correct steps are removed, respectively, as indicated by positive effect sizes (0.33 and 0.19).
\begin{table}[H]
\centering
\caption{This table displays (effect size $d$, $p$) tuples obtained from randomized statistical tests for each type of context variation and each LLM. Larger effect sizes correspond to more LLM sensitivity to ITS context information after adjusting for random chance.}
\begin{threeparttable}
{\setlength{\tabcolsep}{4pt}
\renewcommand{\arraystretch}{1.1} 
\begin{tabular}{c|ccccc}
\hline
\begin{tabular}[c]{@{}c@{}}Effect size,\\ p-value\end{tabular} &
  \begin{tabular}[c]{@{}c@{}}Correct\\ steps\end{tabular} &
  \begin{tabular}[c]{@{}c@{}}Incorrect\\ steps\end{tabular} &
  \begin{tabular}[c]{@{}c@{}}Next\\ steps\end{tabular} &
  Hints &
  \begin{tabular}[c]{@{}c@{}}Knowledge\\ components\end{tabular} \\ \hline
Llama3-8B  & -1.86, .997 & -1.21, .904 & -0.75, .775 & -1.97, .999 & -2.00, .998 \\
Llama3-70B & 0.19, .304 & 2.36, .035\tnote{*} & -1.39, .997 & -1.88, .999 & -1.37, .994 \\
GPT-4o     & -1.66, .995 & 0.33, .293 & -1.68, .999 & -2.16, .999 & -1.90, .998 \\ \hline
\end{tabular}}
\begin{tablenotes}
  \item[*] Significant at the $\alpha = 0.05$ level.
\end{tablenotes}
\end{threeparttable}
\label{tab:p-values}
\end{table}

\subsection{RQ2: Are LLMs Pedagogically Sound?}
\label{sec:results-rq2}

Results regarding the pedagogical quality of LLM responses are summarized in Table \ref{tab:ratings}. Since all metrics are binary quality checks (pass/no pass), we report 95\% confidence intervals for the proportions of successful outcomes.

\begin{table}[H]
\centering
\setlength{\tabcolsep}{6pt} 
\renewcommand{\arraystretch}{1.1} 
\caption{Cross-model comparison for model response pedagogical quality and instruction-following ability. Point estimates (midpoints of 95\% confidence intervals) are shown with their corresponding margins of error.}
\begin{tabular}{c|ccc}
\hline
Metric/model                & Llama3-8B               & Llama3-70B              & GPT-4o               \\ \hline
Resp. to error rating       & $68.25\%\pm13.68\%$     & $47.37\%\pm11.24\%$     & $55.28\%\pm11.19\%$   \\
Praise rating            & $78.62\%\pm8.05\%$      & $68.85\%\pm7.30\%$      & $66.26\%\pm7.39\%$    \\
Intension inclusion         & $92.49\%\pm5.42\%$      & $95.03\%\pm4.24\%$      & $97.57\%\pm2.44\%$    \\
Delimiter existence         & $41.76\%\pm10.88\%$     & $95.03\%\pm4.24\%$      & $97.57\%\pm2.44\%$    \\
Recomm. count               & $35.42\%\pm10.54\%$     & $93.76\%\pm4.87\%$      & $97.57\%\pm2.44\%$    \\ \hline
\end{tabular}
\label{tab:ratings}
\end{table}

Overall, the smallest model, Llama3-8B, receives the highest rating for pedagogical quality, while Llama3-70B and GPT-4o achieve lower scores. However, Llama3-8B frequently fails formatting checks, with common issues including (1) omitting the required intention clause (e.g., "[Encourage]"), (2) incorrect delimiter use (\#), and (3) generating only one recommendation instead of three. In contrast, Llama3-70B and GPT-4o exhibit greater formatting reliability.

\subsection{RQ3: Diversity and Type of LLM Instructional Moves}

\begin{figure}[htpb]
    \centering
    \includegraphics[width=0.9\textwidth]{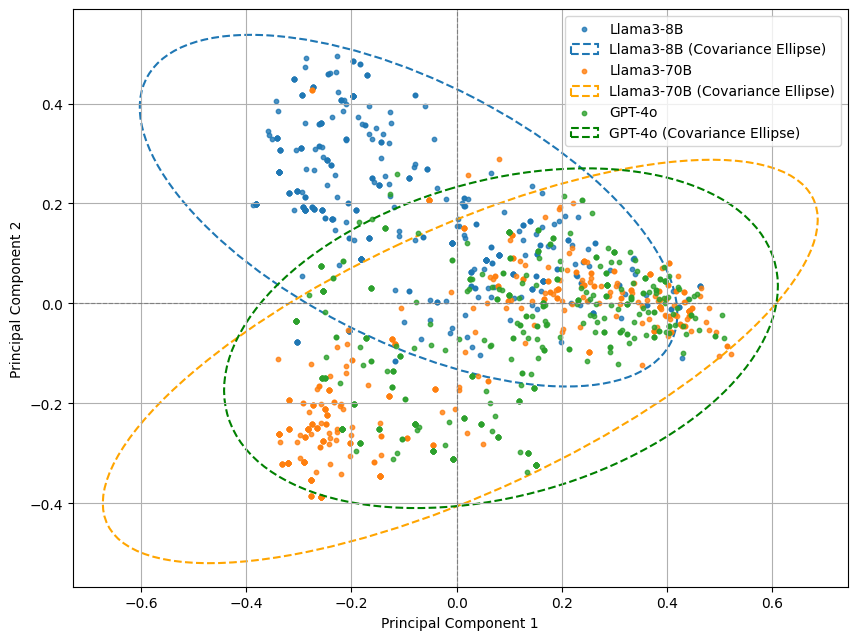}
    \caption{PCA-reduced embedding data colored by backbone LLM}
    \label{fig:pca_scatter}
\end{figure}

We applied PCA (Section \ref{sec:qualitative-analysis}) to reduce the dimensionality of LLM-generated embeddings and visualize them (Fig. \ref{fig:pca_scatter}). The 2D projection includes ellipses representing group covariances, with color-coded groups corresponding to the LLMs, illustrating semantic variation. Notably, Llama3-8 B's embeddings center in the top left, whereas Llama-70B and GPT-4o exhibit substantial overlap.

To better understand the distinctions among these clusters, we informally curated generations. We provide two sets of examples to illustrate these differences. The first concerns fluency and instruction adherence. While Llama3-80B and GPT-4o consistently follow instructions (Table \ref{tab:ratings}) and produce readable text, Llama3-8B generates the following three examples.

\begin{mdframed}[linewidth=1pt]
Example 1: \textit{"[vala…the… [sic Horton but but but ..."} ("but" repeats) \newline
Example 2: \textit{"[Ask to self-explain] Tell me what you're thinking about this problem. What do you think we should do to solve for x?"} \newline
Example 3: \textit{"I appreciate your effort so far! Tell me what you think you should do next with the equation 3x-1=8. \# Talk about it some more \# Great job on simplifying the left side of the equation! \#"}
\end{mdframed}

Example 1 demonstrates garbled text generation \cite{holtzman2020curiouscaseneuraltext} and repetitive output issues \cite{welleck2019neuraltextgenerationunlikelihood}, which are prevalent in smaller language models. Models like Llama3-8B are more likely to produce incoherent text and repeating outputs than larger counterparts. Example 2 highlights Llama3-8 B's inconsistency in generating exactly three recommendations, corroborated by Table \ref{tab:ratings}. Example 3 omits intentions enclosed in [brackets]. Larger models excel at following instructions and often offer more specific and detailed guidance. The example below presents recommendations from three models when given the same problem-solving context.

\begin{mdframed}[linewidth=1pt]
Situation: The child solver is unsure how to remove the constant from the equation, views multiple hints during this step, and asks for the parent's assistance. The current step the student is on is 4x - 2 = 6.\newline
Llama3-8B: \textit{"[Ask to self-explain] Tell me what you think you should do next to solve for x."} \newline
Llama3-70B: \textit{"[Ask to self-explain] I like how you're thinking ahead, can you explain why you think you should add or subtract the 2?"} \newline
GPT-4o: \textit{"[Provide guidance] Remember the concept of canceling constants; adding or subtracting can help isolate the term with x before you divide."}
\end{mdframed}

Notably, Llama3-8B provides only generic guidance lacking scenario specificity, whereas the other two models incorporate the current step and the associated KC. GPT-4o exhibits the highest specificity without inadvertently revealing the correct next step, a flaw observed in Llama3-70B. This variation in attention to detail also contributes to the clustering differences in Fig. \ref{fig:pca_scatter}. A broader trend is that GPT-4o communicates in a more direct tone, whereas Llama models tend to offer higher-level guidance with a somewhat softer approach:

\begin{mdframed}[linewidth=1pt]
Situation: The child is given the problem 2x+1=5x+10 and has not made an attempt. The child asks the parent about dividing both sides with 2 or 5 to remove the coefficient of x's, which is the wrong direction.\newline
Llama3-8B: \textit{"[Ask to self-explain] Tell me what you're thinking about solving this equation. What's your plan so far?"} \newline
Llama3-70B: \textit{"[Encourage start] Let's start by using the equation 2x + 1 = 5x + 10. What's the first step you think we should take to solve for x?"} \newline
GPT-4o: \textit{"[Ask to self-explain] Why do you think we divide the number from both sides?"}
\end{mdframed}

GPT-4o's feedback was comparatively more direct in this example. It explicitly prompts the student about a specific step rather than first assessing their understanding and posing a higher-level question about their approach. The latter would be more appropriate, as neither of the steps mentioned by the student (division by 2 or 5) would be valid, as the ITS permits only whole-number division. This approach deviates from effective tutoring, which encourages open-ended questioning \cite{thomas2023tutor,lin2024can}. We observed this issue in other cases, aligning with GPT-4o's relatively low rating by the tutor training classifier in Table \ref{tab:ratings}.

\section{Discussion}

LLMs enable dialog-based instruction but have been argued to lack the pedagogy of ITS. We examined if LLMs can replicate ITS adaptivity through benchmarking LLM instructional moves. We developed a prompt variation framework that systematically removed key tutoring context elements and tested Llama3-8B, Llama3-70B, and GPT-4o. We assessed adaptivity using text embeddings and randomization tests on 1,350 moves. Classifiers evaluated pedagogical soundness.

\subsection{Discussion of Key Findings}

Addressing \textbf{RQ1} related to whether LLMs can reproduce typical ITS adaptivity in real-world tutoring scenarios, our results suggest that, surprisingly, most LLMs exhibited minimal adaptivity. Only Llama3-70B demonstrated statistically significant responsiveness to student errors. This is notable given that feedback and scaffolding based on accuracy is integral to ITS effectiveness \cite{vanlehn2006behavior,koedinger2012knowledge}. The lack of adaptivity to other critical context elements, such as knowledge components and hints, further underscores the gap between LLMs and ITS adaptivity.

Regarding \textbf{RQ2}, examining if LLMs generate pedagogically desirable responses, the analysis using validated tutor-training classifier \cite{thomas2024tutors} revealed notable model differences. While Llama3-8B received the highest pedagogical soundness ratings, it often failed to follow formatting instructions, making it unreliable for deployment. GPT-4o, in contrast, demonstrated strong instruction-following behavior but tended to provide overly direct feedback, contradicting effective instructional principles \cite{thomas2023tutor,lin2024can,venugopalan2024combining}. These results align with prior studies noting that LLMs' instructional coherence and effectiveness are inconsistent \cite{liffiton2023codehelp,stamper2024enhancing}.

For \textbf{RQ3}, we qualitatively analyzed the diversity of instructional moves generated by LLMs. Findings reveal that larger models generally provide more detailed and specific guidance than smaller ones. GPT-4o, for instance, delivers precise but often overly direct feedback, misaligning with best tutoring practices \cite{thomas2023tutor,lin2024can}. In contrast, Llama3-8B produced more open-ended responses but often failed to align recommendations with relevant problem-solving steps. This underscores a tradeoff between generality and specificity in LLMs, affecting their suitability for tutoring. Balancing the specificity of instructional support—the assistance dilemma \cite{koedinger2007exploring}—is a fundamental AIED design issue. Our findings suggest that seemingly minor choices, such as model selection, influence the degree of assistance provided under identical prompts. Thus, effective LLM-based tutoring requires tuning parameters like model temperature \cite{agarwal2024understanding} to optimize scaffolding balance. Future research may systematically explore the effect of tuning these parameters on instructional quality using our benchmarking method.

\subsection{Implications for LLM-Based Tutoring}

Our findings contribute to the ongoing debate on the viability of LLMs as tutoring agents, increasingly adopted in AIED environments \cite{shetye2024evaluation,schmucker2023ruffle,venugopalan2024combining}. While prior work suggests LLM-generated hints can yield learning gains comparable to expert-authored hints \cite{pardos2024chatgpt}, tutoring effectiveness extends beyond hint provision. Meta-analyses show that ITS instruction outperforms standard curricula in improving learning outcomes \cite{kulik2016effectiveness,steenbergen2013meta}, leveraging multiple adaptive dimensions (e.g., hints, feedback, problem selection \cite{kulik2016effectiveness}). As even the best-performing LLM in our study only marginally approximated ITS adaptivity, our results suggest LLM-based tutoring is unlikely to match ITS learning benefits without improvements on benchmarks like ours, which researchers can build on. Moreover, concerns persist that students may use LLMs in ways that reduce cognitive effort \cite{fan2024beware}. Hence, future research may prioritize hybrid settings that embed LLMs within ITS frameworks \cite{venugopalan2024combining} rather than seeking to replace ITS.

\subsection{Limitations and Future Work}

First, our benchmarking study examines a single tutoring system within a specific instructional domain and a limited sample size (algebraic equation solving). While this allows for a controlled analysis of adaptivity, the findings may not generalize to other educational settings, such as open-ended problem-solving or non-mathematical subjects. Future research could apply our open-source benchmarking approach across larger data sets and diverse disciplines. Second, the impact of context window length in LLM-based tutoring remains an open question. Our study provided full student attempt histories, but selecting targeted subsets of context data may enhance LLM generations—explorations beyond the present study's scope. Third, our findings may be limited by using tutoring scenarios from an American sample encoded in English, potentially affecting LLM performance in languages underrepresented in web training corpora. Future research could expand our benchmarking approach to multilingual data sets.

\section{Conclusion}

We contribute a novel and open-source benchmarking method to assess whether large language models (LLMs) can replicate intelligent tutoring systems (ITS) adaptivity with high instructional fidelity. Our findings indicate that current LLMs struggle to respond effectively to key context signals, such as student errors and knowledge components, essential for ITS adaptivity. While Llama3-70B demonstrated some sensitivity to student errors, neither it nor GPT-4o consistently aligned instructional moves with pedagogy driving ITS effectiveness. The smaller Llama3-8B model received higher ratings for response quality but frequently failed to follow critical instructions (e.g., output formatting), reducing their reliability for real-time tutoring. These results highlight that LLM-based tutoring still lacks the structured, context-driven support that defines ITS. Despite their linguistic fluency, LLMs require significant improvement in delivering nuanced, pedagogically sound scaffolding. While LLMs show promise for conversational learner support, precise methods are needed to match established tutoring systems' adaptive rigor and instructional quality. We conclude that LLMs are, at present, unlikely to produce learning benefits similar to those widely documented for intelligent tutoring.

\section*{Acknowledgements}

This research was funded by the Institute of Education Sciences (IES) of the U.S. Department of Education (Award \#R305A220386).

\bibliographystyle{splncs04}
\bibliography{main} %

\end{document}